\documentclass[runningheads]{llncs}
\usepackage[T1]{fontenc}
\usepackage{graphicx}
\usepackage{amsmath}
\usepackage{booktabs}
\usepackage[misc]{ifsym}
\newcommand{\corr}{(\Letter)}
\usepackage[nolist]{acronym}
\usepackage{hyperref}

\begin{document}

\title{An Optimization Framework for Automated Assessment of Biological Plausibility of Spiking Neurons}

\titlerunning{Automated Assessment of Biological Plausibility of Spiking Neurons}

\author{
    Sven Nitzsche\inst{1} \corr \and 
    Alexandru Ionita\inst{2} \and
    Andreas Faust\inst{1} \and
    Bogdan Ionescu\inst{2} \and
    Juergen Becker\inst{3}
}

\authorrunning{S. Nitzsche, A. Ionita et al.}

\institute{
    FZI Research Center for Information Technology, Karlsruhe, Germany \email{nitzsche@fzi.de}
    \and
    AI Multimedia Lab, CAMPUS Research Institute, National University of Science and Technology Politehnica Bucharest, Romania \email{\{aionita1204, bogdan.ionescu\}@upb.ro}
    \and
    Karlsruhe Institute of Technology, Karlsruhe, Germany
}

\maketitle

\begin{abstract}
Biological plausibility is a key concept in neuromorphic computing and spiking neural networks, yet it remains inconsistently defined and difficult to quantify. In this work, we present an open-source framework for the automated assessment of biological plausibility in spiking neuron models. Our method builds on the idea of evaluating a model’s ability to replicate canonical neuronal firing patterns observed in biological systems, following the classification proposed by Izhikevich. 
By encoding these patterns into objective functions and optimizing model parameters accordingly, our framework enables empirical assessment without requiring prior analytical modeling. Treating neuron models as black boxes, it provides a practical and flexible means of characterizing their dynamic capabilities. We demonstrate the effectiveness of the framework on several established models and a previously unexplored custom model. Implemented in Python and compatible with PyTorch and the Norse library, the framework is tailored for machine learning contexts. It is intended as a starting point for systematic research into the relationship between biological plausibility and network-level performance metrics such as accuracy, energy efficiency, robustness, and adaptability.
\end{abstract}

\begin{acronym}
    \acro{snn}[SNN]{Spiking Neural Network}
    \acro{lif}[LIF]{Leaky Integrate-and-Fire}
    \acro{adex}[AdEx]{Adaptive Exponential}
    \acro{dc}[DC]{direct current}
\end{acronym}
\section{Introduction}
\label{sec:introduction}

Neuromorphic computing and \acp{snn} in particular are emerging as promising paradigms in the field of embedded machine learning, largely due to their energy efficiency and advantages in tasks with time-dependent data~\cite{blouw2019benchmarking,stoffel2024spiking,dampfhoffer2022snns}. The fundamental concept driving neuromorphic computing is to emulate biological mechanisms, aiming to leverage the computational efficiency observed in biological systems~\cite{stiefel2023energy}. Contemporary models incorporate several principles derived from neuroscience, such as asynchronous processing, spike-based transmission of information, and dynamic, stateful neuron behavior. Adopting biologically inspired principles seems to be beneficial, not only for modeling cognitive functions in neuroscience contexts~\cite{pulvermuller2021biological}, but also in practical machine learning applications~\cite{sanaullah2023exploring,de2023analysis,ganguly2024spike}.

However, assessing the degree of biological plausibility in such models remains challenging. It is still an open question which specific biological features are necessary or beneficial for particular tasks~\cite{almog2016realistic}. In addition, the term 'biological plausibility' is frequently used inconsistently, often relying on subjective judgment rather than quantifiable criteria. This results in ambiguous claims of plausibility that lack empirical support and are devoid of any explicit correlation with quantifiable neural phenomena~\cite{love2021levels}. 
To address these issues, there is a need for standardized, quantifiable metrics of biological plausibility to enable more rigorous and objective benchmarking. Particularly for neuron models, there already are multiple definitions of this concept, broadly classified into analytical and empirical approaches, each with unique advantages and disadvantages.

\textbf{Our Contribution:}
In this work, we introduce an open-source framework designed to automatically assess the biological plausibility of arbitrary neuron models\footnote{\url{https://github.com/nitzsche-fzi/Bioplaus}}. Our approach aims to integrate the strengths of both analytical and empirical strategies, offering a generalizable way to evaluate biological realism without requiring extensive mathematical analysis of neuronal dynamics. The framework is implemented in Python and is fully compatible with PyTorch as well as SNN libraries such as Norse~\cite{norse2021}, facilitating its adoption in both research and productive environments.

The outline of this paper is as follows. Section~\ref{sec:background} provides background information on neuron models and firing patterns used in this work. It also reviews definitions of biological plausibility of neuron models in literature. Section~\ref{sec:opt_framework} details our framework, including objective functions and optimization algorithms. Section~\ref{sec:evaluation} assesses the framework on four neuron models, including a custom model previously unexplored in literature. Finally, Section~\ref{sec:discussion} summarizes the framework’s contributions, puts them in a larger context and highlights future directions for improvement.
\section{Background}
\label{sec:background}

\begin{figure}[t]
    \centering
    \includegraphics[width=\linewidth]{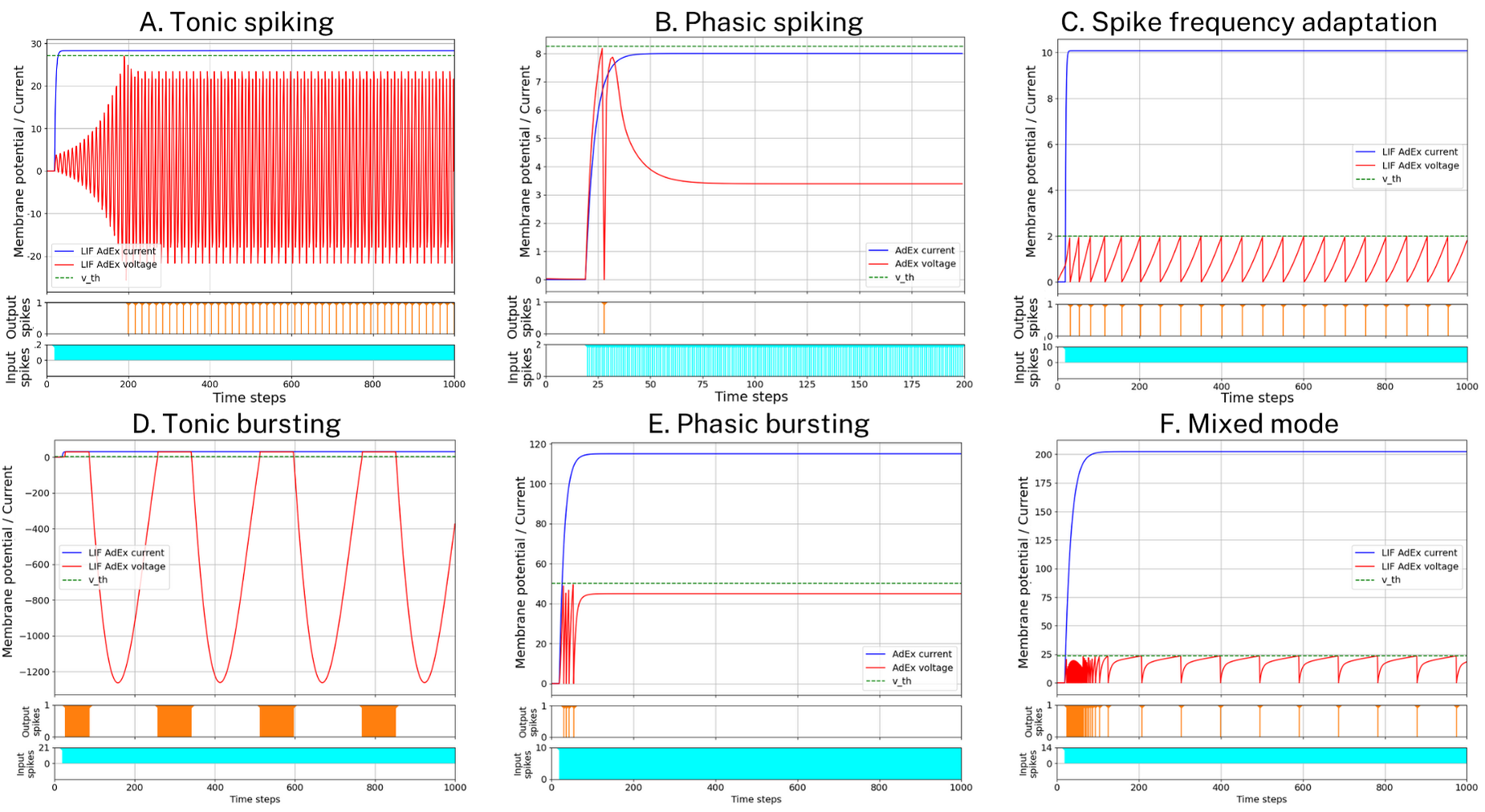}
    \caption{Firing patterns for DC input. Dimensionless voltages and currents against time steps in ms. All patterns were automatically generated with the optimization framework from the \ac{adex} model.}
    \label{fig:voltages_DC}
\end{figure}

For this work, we focus on the biological realism of the dynamics of single neurons, excluding aspects such as network connectivity patterns or dendritic computation.
Neuron models form the computational core of \acp{snn}, governing how inputs are integrated into spikes.
The development of neuron models began with early electrophysiological experiments on how neurons respond to electrical stimulation. These experiments revealed key principles of neuronal excitability and firing behavior, forming the basis for modeling neurons as electrical systems that describe how they integrate and transmit signals under different conditions~\cite{gerstner2014neuronal}.


\subsection{Biological Plausibility}
\label{subsec:bioplausibility}
Multiple attempts have been made to define criteria for the biological plausibility of neuron models. Izhikevich defines plausibility through a neuron’s ability to reproduce 20 canonical neuro-computational features observed in biological neurons, focusing on their internal and output behavior for given inputs~\cite{izhikevich2004whichmodel}. This approach prioritizes functional equivalence over biophysical detail. It requires a sound understanding of a neuron's equations to estimate which of the 20 features are possible to achieve with a given model and thorough testing to find matching neuron parameters.

As an alternative to this analytical approach, neuron models can be fitted directly to experimental data recorded from actual biological neurons to make them as realistic as possible~\cite{druckmann2007novel,marin2020optimization,gerstner2009good}. Following this approach, Jolivet et al. established an empirical framework to assess biological plausibility in the form of a yearly challenge~\cite{jolivet2008quantitative}. Participants predicted various features of cortical neurons, such as spike timings and spike frequency, with performance measured against in vitro and in vivo datasets. While this approach is flexible and applicable to arbitrary neuron models without prior knowledge of their capabilities, it assesses them only in specific scenarios. Consequently, it aims to identify the most plausible parameter set. This may lead to overfitting, possibly reducing the universality of the assessed bioplausibility, compared to more analytical methods.

\subsection{Neuron Models}
\label{subsec:neuron_models}

\begin{figure}[t]
    \centering
    \includegraphics[width=\linewidth]{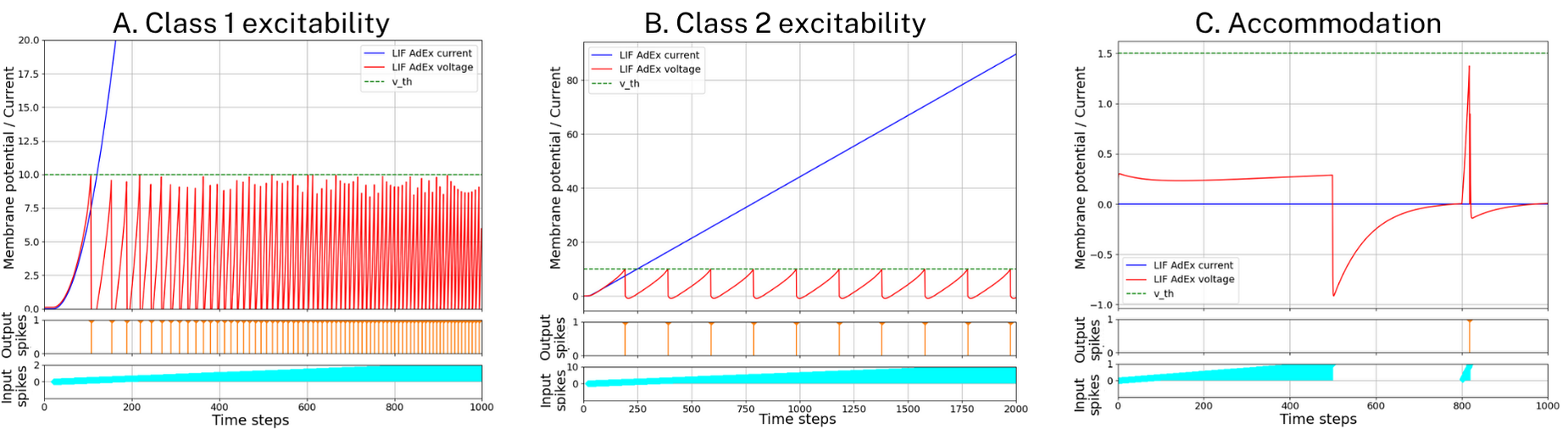}
    \caption{Firing patterns for linear input. Dimensionless voltages and currents against time steps in ms. All patterns were automatically generated with the optimization framework from the \ac{adex} model.}
    \label{fig:voltages_linear}
\end{figure}

Within this work, we use four neuron models. The \ac{lif}, \ac{adex}, and Izhikevich models have already been extensively studied in literature \cite{izhikevich2004whichmodel,gerstner2014neuronal,norse2021}, thus we only provide a short summary here. For an overview of the equations, please see the original papers or our Git repository.

\textbf{The \ac{lif} model} is a simple neuronal representation, often described by an RC-circuit analogy. It is computationally efficient and widely used as an activation function in \acp{snn}. Its simplicity allows for analytical solutions and large-scale simulations, but at the cost of reduced biological plausibility.

\textbf{The \ac{adex} model} extends the \ac{lif} framework by adding an exponential term to capture nonlinear spike initiation and an adaptation current to reflect spike-frequency adaptation, resulting in greater biological realism. However, the exponential nonlinearity also increases computational demands.

\textbf{The Izhikevich model} strikes a balance by using simple yet versatile equations to reproduce a range of neuronal firing patterns. Although phenomenological, it captures essential neuronal dynamics at a low computational cost by employing a quadratic nonlinearity instead of the more expensive exponential term in \ac{adex}.

\begin{figure}[t]
    \centering
    \includegraphics[width=\linewidth]{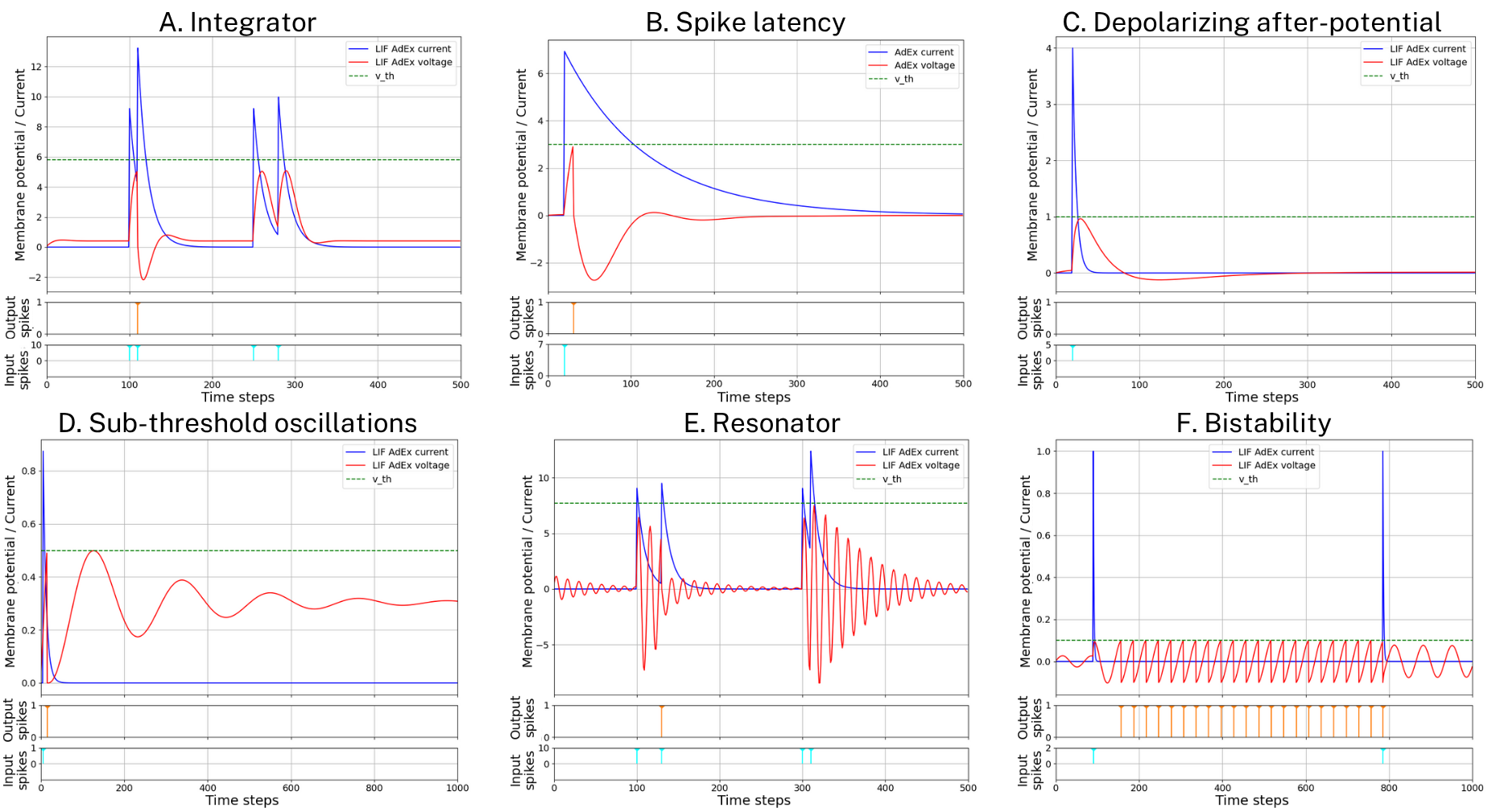}
    \caption{Firing patterns for spiked input. Dimensionless voltages and currents against time steps in ms. All patterns were automatically generated with the optimization framework from the \ac{adex} model.}
    \label{fig:voltages_spikes}
\end{figure}

\textbf{The N2D2 model} is our custom-made neuron model we use internally for other research. Here, it demonstrates the capability of the proposed framework in assessing the bioplausibility of a model that has not been characterized regarding its biological realism. It is defined by the following differential equations:
\begin{align*}
    \dot{v} &= a v + b i u + c u^2\\
    \dot{u} &= d u + e i^2
\end{align*}
together with the jump condition
\begin{align*}
    z &= \Theta(v - v_{\text{th}}) \\
    v &= (1-z) v + z (f i + g u) \\
    u &= (1-z) u + z (h i + j u),
\end{align*}
where $a$ to $h$ and $j$ are model parameters. The second potential $u$ integrates over the squared input $i$, with leakage controlled by the $d$ parameter. The membrane potential $v$ depends on both $u$ and $i$, implementing a complex threshold that is difficult to interpret biologically.
It is a non-leaky integrator of the squared input value in its second internal state, with the first state serving as a complex thresholding mechanism that depends on the square of the internal state and the product of the input current and the second internal state. Adaptive thresholds are a common feature in biological neurons, but the complexity of the thresholding mechanism in this model is hard to interpret biologically, leaving the biological plausibility of the model in question.

\begin{figure}[t]
    \centering
    \includegraphics[width=\linewidth]{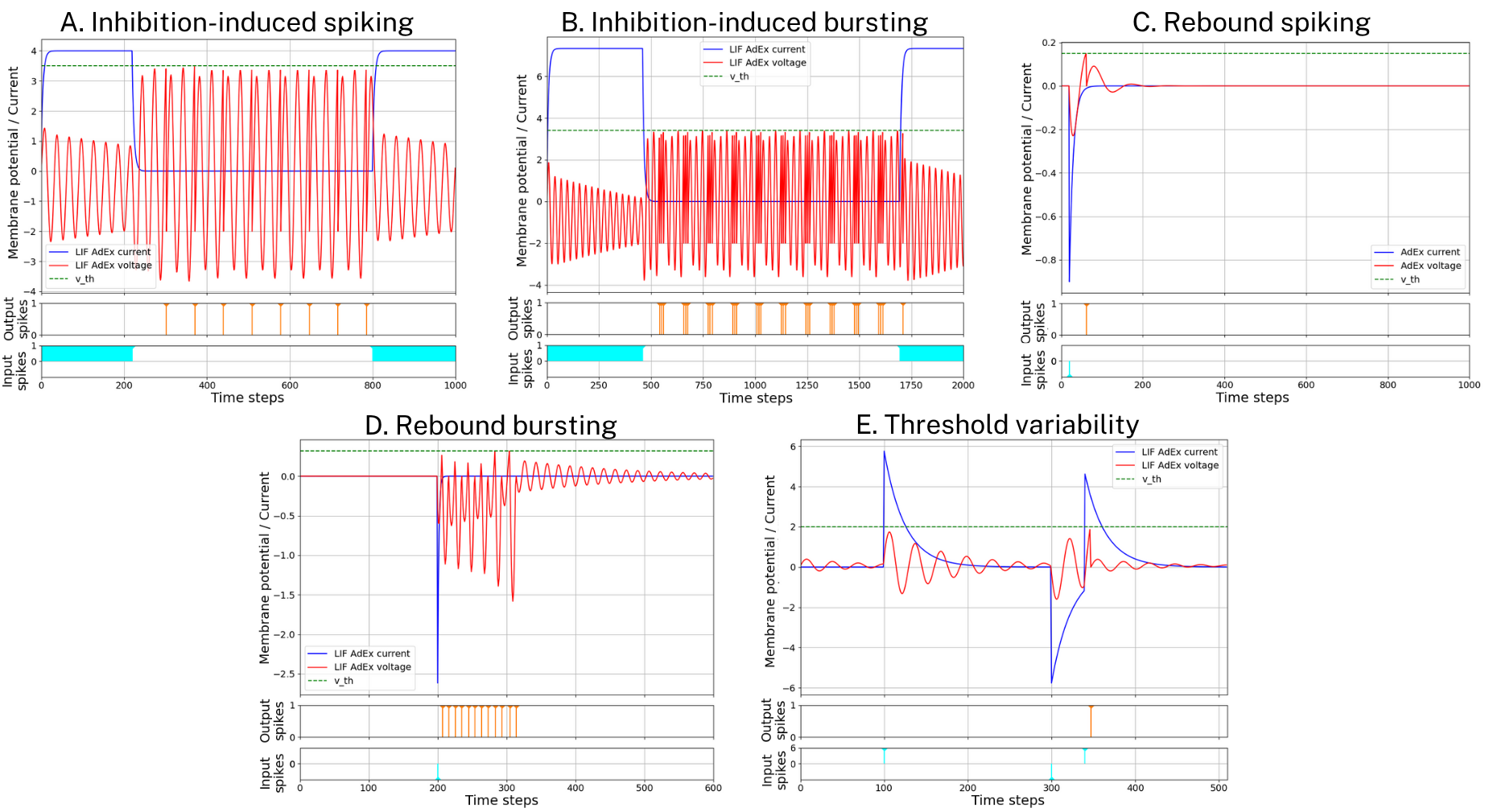}
    \caption{Firing patterns for inhibitory input. Dimensionless voltages and currents against time steps in ms. All patterns were automatically generated with the optimization framework from the \ac{adex} model.}
    \label{fig:voltages_inhibition}
\end{figure}

\subsection{Firing Patterns}
\label{subsec:firing_patterns}

\begin{figure}[t]
    \centering
    \includegraphics[width=\linewidth]{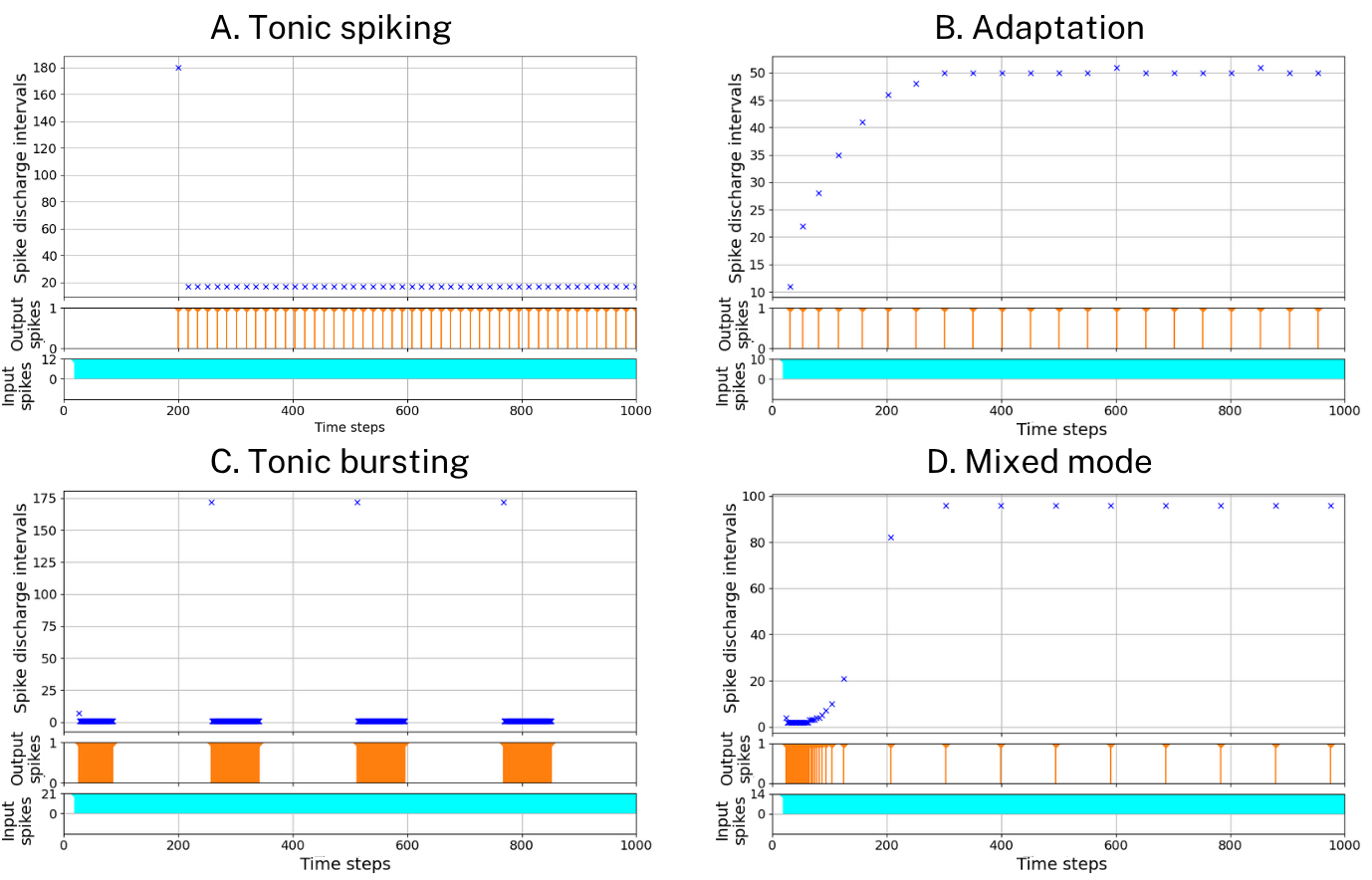}
    \caption{Spike discharge intervals against time steps, both in ms, corresponding to firing patterns with DC input. All patterns were automatically generated with the optimization framework from the \ac{adex} model.}
    \label{fig:discharge_DC}
\end{figure}

As mentioned above, the range of possible firing patterns can be used as an indicator for the bioplausibility of neuron models. For this work, we focus on the relationship between input stimuli and output responses, without considering the underlying equations. The input current is sorted into four main categories: constant direct current (DC IN), linear current (lin. IN), inhibitory current (inh. IN), and spiked current (spiking IN). Similarly, the output firing patterns fall into three distinct types: tonic, phasic, and subthreshold (subth. OUT). All firing patterns illustrated in the indicated plots were simulated using the \ac{adex} model. Following Izhikevich \cite{izhikevich2004whichmodel}, the key firing patterns observed in biological neurons are:\\
\textbf{Tonic spiking} (DC IN, tonic OUT - Figures \ref{fig:voltages_DC}A and \ref{fig:discharge_DC}A): firing spikes at a steady rate. This pattern, observed in cortical and motor neurons, ensures sustained information transmission and serves as a key test for neuron models.\\
\textbf{Phasic spiking} (DC IN, phasic OUT - Figure \ref{fig:voltages_DC}B): firing a single spike before becoming silent, even if the input persists. This pattern is important for detecting transient stimuli and is commonly observed in sensory neurons, where it helps encode changes rather than sustained signals.\\
\textbf{Spike frequency adaptation} (DC IN, tonic OUT - Figures \ref{fig:voltages_DC}C and \ref{fig:discharge_DC}B): the firing rate decays to a tonic spiking steady state. It emphasizes transient changes, thus improving the detection of novel or rapidly changing inputs \cite{ganguly2024spike}.\\
\textbf{Tonic bursting} (DC IN, tonic OUT - Figures \ref{fig:voltages_DC}D and \ref{fig:discharge_DC}C): temporal spike clusters in the output, separated by brief, regular quiescent periods. This enhances signal transmission by combining the reliability of tonic spiking with the additional temporal modulation provided by bursts.\\
\textbf{Phasic bursting} (DC IN, phasic OUT - Figure \ref{fig:voltages_DC}E): similar to phasic spiking. It occurs when a neuron generates a brief burst of spikes in response to a constant DC input, but then becomes quiescent, even if the stimulus continues.\\
\textbf{Mixed mode} (DC IN, tonic OUT - Figures \ref{fig:voltages_DC}F and \ref{fig:discharge_DC}D): a burst at the onset of stimulation, followed by tonic spiking in the steady state.\\
\textbf{Class 1 excitability} (lin. IN, tonic OUT - Figures \ref{fig:voltages_linear}A and \ref{fig:discharge_linear}A): the neuron fires regularly even at small input current, with the firing rate increasing when the input strength increases. Beyond a threshold, the output spike frequency remains constant.\\
\textbf{Class 2 excitability} (lin. IN, tonic OUT - Figures \ref{fig:voltages_linear}B and \ref{fig:discharge_linear}B): the neuron remains silent below a threshold input strength, but once the threshold is reached, it fires at a constant rate, even when the input strength is increased.\\
\textbf{Accommodation} (lin. IN, phasic OUT - Figure \ref{fig:voltages_linear}C): the gradual decrease in a neuron's excitability in response to a sustained stimulus, preventing excessive firing. This is tested by applying two linearly increasing input currents: one that ramps up slowly, and another weaker but more sharply ramped current. In the first case, the neuron should not fire, whereas in the second case it should.\\
\textbf{Integrators} (spiked IN, phasic OUT - Figure \ref{fig:voltages_spikes}A): accumulate input signals over time, gradually raising the membrane potential until it reaches the threshold to trigger a spike.\\
\textbf{Spike latency} (spiked IN, phasic OUT - Figure \ref{fig:voltages_spikes}B): delay between the onset of a stimulus and the neuron’s generation of a spike.\\
\textbf{Depolarizing after-potential} (spiked IN, subth. OUT - Figure \ref{fig:voltages_spikes}C): period of depolarization of the action potential of a neuron after firing a spike, where the membrane potential briefly becomes more positive than its resting state.\\
\textbf{Subthreshold oscillations} (spiked IN, subth. OUT - Figure \ref{fig:voltages_spikes}D): oscillations in a neuron's membrane potential that occur below the threshold. This mechanism potentially unlocks other firing patterns, such as resonance or bistability.\\
\textbf{Resonators} (spiked IN, phasic OUT - Figure \ref{fig:voltages_spikes}E): when stimulated with pulsed inputs, a resonator neuron can amplify signals that match its natural oscillation frequency, effectively "resonating" with the input.\\
\textbf{Bistability} (spiked IN, tonic OUT - Figures \ref{fig:voltages_spikes}F and \ref{fig:discharge_spikes}A): a neuron's ability to exist in two stable states: a resting state and an active firing state. We switch between these states with an input pulse, whose timing must be synchronized with the neuron's subthreshold oscillations.\\
\textbf{Inhibition-induced spiking} (inh. IN, tonic OUT - Figures \ref{fig:voltages_inhibition}A and \ref{fig:discharge_inhibition}A): a normally inactive neuron fires at a constant rate in response to an inhibitory input.\\
\textbf{Inhibition-induced bursting} (inh. IN, tonic OUT - Figures \ref{fig:voltages_inhibition}B and \ref{fig:discharge_inhibition}B): produces bursts of multiple spikes upon the release of inhibition.\\
\textbf{Rebound spiking} (inh. IN, phasic OUT - Figure \ref{fig:voltages_inhibition}C): fires a spike immediately after removing an inhibitory input.\\
\textbf{Rebound bursting} (inh. IN, phasic OUT - Figure \ref{fig:voltages_inhibition}D): produces a burst of spikes upon the release of inhibitory input, due to a strong depolarization.\\
\textbf{Threshold variability} (inh. IN, phasic OUT - Figure \ref{fig:voltages_inhibition}E): fluctuations in a neuron's spiking threshold, influenced by prior activity. A positive input pulse fails to trigger a spike on its own but successfully induces firing when preceded by an inhibitory input, indicating a temporary lowering of the threshold.

The \ac{lif} model primarily exhibits tonic spiking, integrating inputs over time but lacking the ability to capture more intricate biological patterns such as adaptation or bursting. The \ac{adex} model can generate all the patterns discussed above, offering a more biologically realistic representation of neuronal activity, however, with higher computational costs. Similarly, the Izhikevich model also reproduces all the observed patterns but with better computational efficiency.
\section{Optimization Framework}
\label{sec:opt_framework}

Evaluating how many biologically observed firing patterns a neuron model can reproduce offers a practical and objective measure of bioplausibility and might be more universal than the empirical approach described in Section~\ref{subsec:bioplausibility}. Hence, we use this as base for our work.
However, manual parameter tuning is time-consuming and inefficient. To address this, we propose a semi-automated, empirical framework for evaluating the bioplausibility of neuron models, tailored to machine learning applications.

The core idea is to define a specific objective function for each desired firing pattern, constructed such that its minimized when a given neuron model exhibits that pattern. We thus reformulate the task into a combinatorial optimization problem and discover firing patterns automatically using suitable optimization algorithms. 
For each pattern, the algorithm will attempt to find one matching set of parameters. 
Section \ref{subsec:framework_outline} outlines the general workflow of the proposed approach. This is followed by Section \ref{subsec:objective_function}, which provides a detailed discussion on the design of objective functions, and Section \ref{subsec:optimization_algorithm}, which addresses the selection of appropriate optimization algorithms for efficiently identifying these patterns.

\subsection{Framework Outline}
\label{subsec:framework_outline}

\begin{figure}[t]
    \centering
    \includegraphics[width=\linewidth]{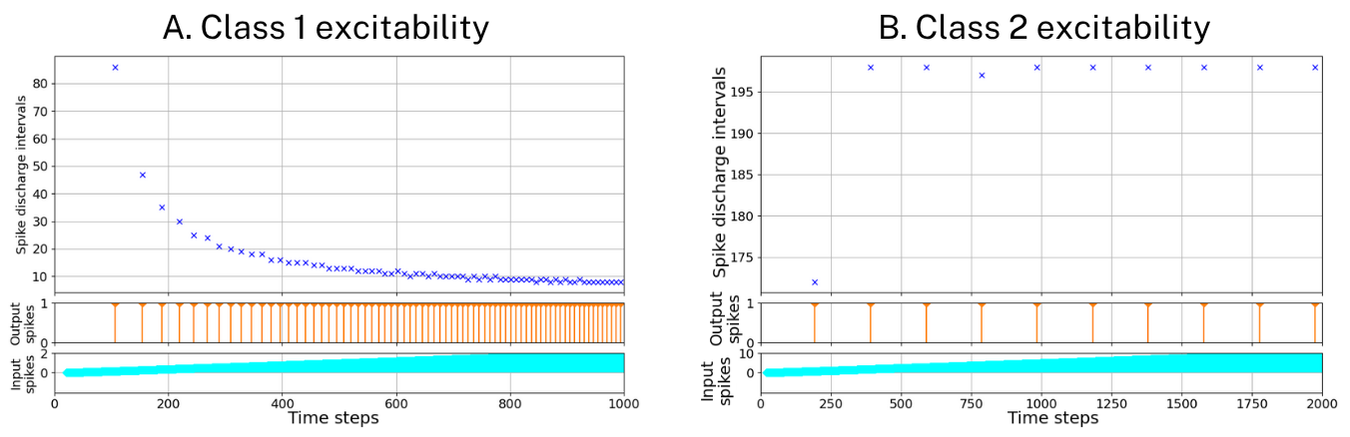}
    \caption{Spike discharge intervals against time steps, both in ms, corresponding to firing patterns with linear input. All patterns were automatically generated with the optimization framework from the \ac{adex} model.}
    \label{fig:discharge_linear}
\end{figure}

We designed the framework for practical integration into a typical \ac{snn} design workflow. It is implemented in Python and compatible with spiking neuron models based on the Norse library \cite{norse2021}, supporting both built-in and custom neuron models in a model-agnostic manner. The main steps of the proposed bioplausibility assessment are outlined below.

\subsubsection{Initial Model Exploration.} An initial investigation of the neuron model is necessary to identify relevant parameter ranges. While the framework theoretically treats models as black boxes, leaving parameters unrestricted increases the search space, making convergence difficult. A preliminary understanding, such as parameter scaling, typical value ranges, or eliminating irrelevant parameters, can significantly improve efficiency while remaining far less demanding than a full analytical study. Some firing patterns are more elusive and require optimization over input variations in addition to model parameter sweeps, which can also be identified during this initial stage. For instance, in the case of bistability, input timing plays a critical role.

\subsubsection{Automated Optimization.} We encode each pattern into a quantitative objective function, designed to reach its minimum only when the corresponding pattern is present. For instance, tonic spiking can be captured by minimizing the variance of inter-spike intervals - zero variance indicates spiking at a constant rate. To exclude undesired behaviors that may also minimize the function (e.g., spiking at every timestep), we introduce penalty terms, such as penalizing discharge intervals below a threshold. These objective functions are then integrated into an optimization framework. We implemented our framework on top of the Optuna package \cite{optuna_2019}, which offers a flexible and robust platform with a wide selection of pre-implemented optimization algorithms, as well as straightforward support for incorporating custom algorithms into the workflow. We discuss some details of objective function design and the choice of optimization algorithms in subsequent sections.

\subsubsection{Post-Optimization Validation.} Once we find an optimum, the solution must be validated to ensure that the desired firing pattern has indeed been achieved. If the pattern is not correctly recovered despite a minimized objective function, this typically means it is either not supported by the neuron model or parameter search space is to wide and should be narrowed. In some cases, manually adjusting the solution found by the optimizer can further enhance the result.

\subsection{Objective Functions}
\label{subsec:objective_function}

\begin{figure}[t]
    \centering
    \includegraphics[width=0.5\linewidth]{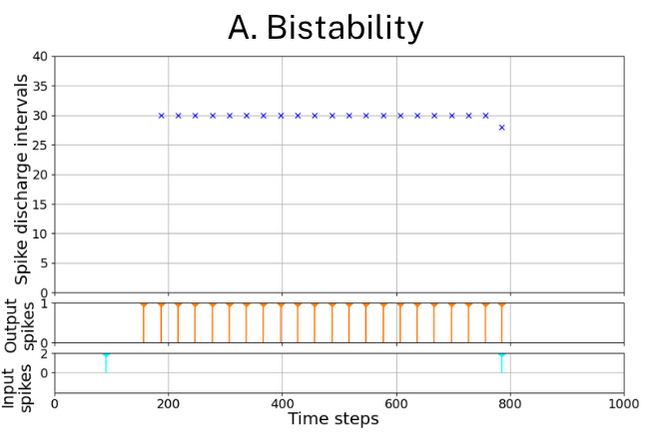}
    \caption{Spike discharge intervals against time steps, both in ms, corresponding to firing patterns with spiking input. All patterns were automatically generated with the optimization framework from the \ac{adex} model.}
    \label{fig:discharge_spikes}
\end{figure}

\begin{figure}[t]
    \centering
    \includegraphics[width=\linewidth]{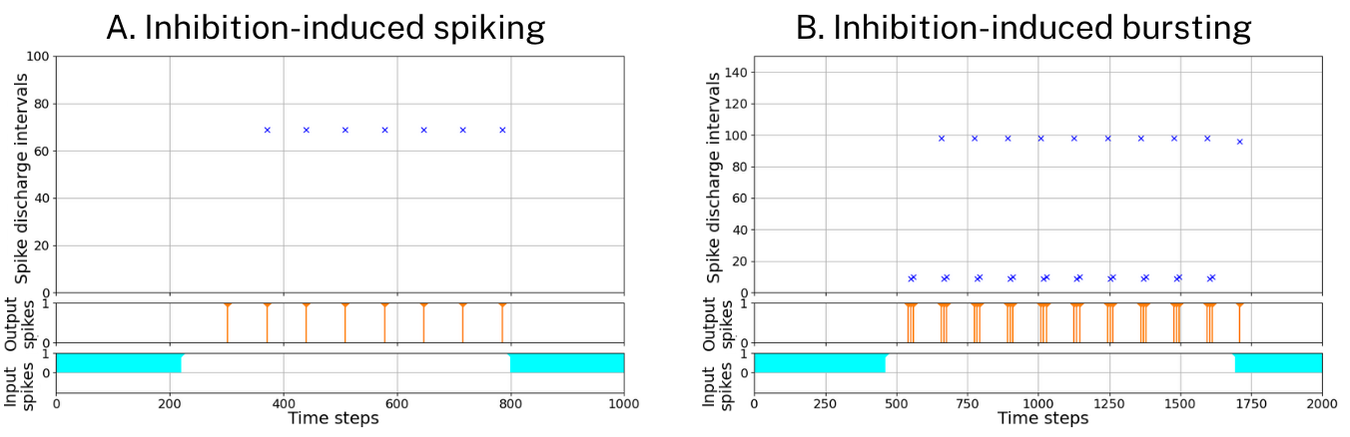}
    \caption{Spike discharge intervals against time steps, both in ms, corresponding to firing patterns with inhibitory input. All patterns were automatically generated with the optimization framework from the \ac{adex} model.}
    \label{fig:discharge_inhibition}
\end{figure}

The main features when formulating optimization objectives are the delay of the first output spike and the sequence of time intervals between consecutive spike discharges. The first-spike delay is essential for identifying phasic spiking patterns and is also used in general penalty terms, such as penalizing models that spike before the input onset. The spike discharge intervals are central to characterizing patterns involving multiple output spikes and are illustrated in Figures \ref{fig:discharge_DC}, \ref{fig:discharge_linear}, \ref{fig:discharge_spikes}, and \ref{fig:discharge_inhibition}.

This section outlines only the core principles used to define the objectives. For implementation specifics, including the design of penalty terms, please refer to our Git repository.

Tonic spiking (Figures \ref{fig:voltages_DC}A, \ref{fig:discharge_DC}A) is characterized by output spikes being fired with constant frequency under DC input. We define the corresponding objective function as the variance of the spike discharge intervals array, scaled by the number of output spikes. The variance is zero at minimum, when the output spikes are equally spaced. We use the same type of objective for class 2 excitability (Figures \ref{fig:voltages_linear}B, \ref{fig:discharge_linear}B), bistability (Figures \ref{fig:voltages_spikes}F, \ref{fig:discharge_spikes}A) and inhibition-induced spiking (Figures \ref{fig:voltages_inhibition}A, \ref{fig:discharge_inhibition}A). Note that there is no confusion in using the same objective for the above-mentioned patterns, as the input differs.

An alternative tonic spiking objective computes the Manhattan distance between the neuron's output spikes and a template time series, constructed to be identical to the neuron's output given it fires at a constant rate. We assume the spiking frequency of the template is the largest discharge interval in the neuron's output, and choose the offset such that the first spike in the template and the second spike fired by the neuron are synchronized. The template construction can be generalized beyond firing single spikes at a constant rate to firing bursts at a constant rate, where the template of the burst unit is extracted from the neuron's output under the assumption that tonic bursting is indeed present. We use this formulation to automatically assess two other firing patterns: tonic bursting (Figures \ref{fig:voltages_DC}D, \ref{fig:discharge_DC}C) and inhibition-induced bursting (Figures \ref{fig:voltages_inhibition}B, \ref{fig:discharge_inhibition}B).

Spike frequency adaptation (Figures \ref{fig:voltages_DC}C, \ref{fig:discharge_DC}B), mixed mode (Figures \ref{fig:voltages_DC}F, \ref{fig:discharge_DC}D), and class 1 excitability (Figures \ref{fig:voltages_linear}A, \ref{fig:discharge_linear}A) all have a unique behavior at the beginning of the excitation, with uniformly distributed spikes in the steady state. We construct their objectives as a sum of the tonic spiking objective function applied towards the end of the time series and a second term, which only depends on the neuron's output at the beginning of excitation. For adaptation, the discharge intervals gradually increase towards the asymptotic tonic intervals. In mixed mode, the output spikes are agglomerated in the beginning at very small time intervals and then sharply increase towards the asymptotic behavior. In the case of class 1 excitability, the discharge intervals gradually decrease.

Another important behavior is phasic spiking (Figure \ref{fig:voltages_DC}B), where the objective function is the absolute difference between the number of output spikes and one, plus a penalty term for the delay compared to the start of the excitation. We find several other firing patterns involving a single output spike using the same design, by only changing the reference of the delay penalty: accommodation (Figure \ref{fig:voltages_linear}C), integrator (Figure \ref{fig:voltages_spikes}A), spike latency (Figure \ref{fig:voltages_spikes}B), resonator (Figure \ref{fig:voltages_spikes}E), rebound spiking (Figure \ref{fig:voltages_inhibition}C) and threshold variability (Figure \ref{fig:voltages_inhibition}E). In the case of spike latency, we reward the delay, instead of penalizing it.

Only a few patterns remain that do not fall into these categories. For phasic bursting (Figure \ref{fig:voltages_DC}E) and rebound bursting (Figure \ref{fig:voltages_inhibition}D), we define the objective as the number of spikes divided by the time coordinate of the last spike, which promotes outputs with a high output spike density at the beginning of excitation. Depolarizing after-potential (Figure \ref{fig:voltages_spikes}C) is measured by the number of voltage humps following an output spike, under the condition that there are no subsequent output spikes. Finally, we quantify the occurrence of subthreshold oscillations (Figure \ref{fig:voltages_spikes}D) by the number of times the voltage oscillates without firing.

\subsection{Optimization Algorithm}
\label{subsec:optimization_algorithm}

The choice of optimization algorithm influences both the sampling strategy over the parameter space and the convergence behavior of the optimization process. This selection is non-trivial in the present context, as the objective function landscapes are non-convex. Moreover, conventional gradient-based methods are incompatible due to the discontinuous nature of the neuron models, which include jump conditions that violate differentiability assumptions.

We carried out the majority of the optimization tasks in this work using two of Optuna’s built-in samplers: the \textit{RandomSampler}, which performs uniform random sampling, and the \textit{CmaEsSampler} \cite{nomura2024cmaes}, which implements the Covariance Matrix Adaptation Evolution Strategy (CMA-ES). In addition, we implemented a custom approximated quantum-inspired evolutionary algorithm (QEA) and integrated it into the Optuna framework. This QEA variant serves both as an initial step toward incorporating more advanced evolutionary optimization strategies and as a demonstration of the framework’s extensibility.
\section{Evaluation}
\label{sec:evaluation}

\begin{figure}[t]
    \centering
    \includegraphics[width=0.5\linewidth]{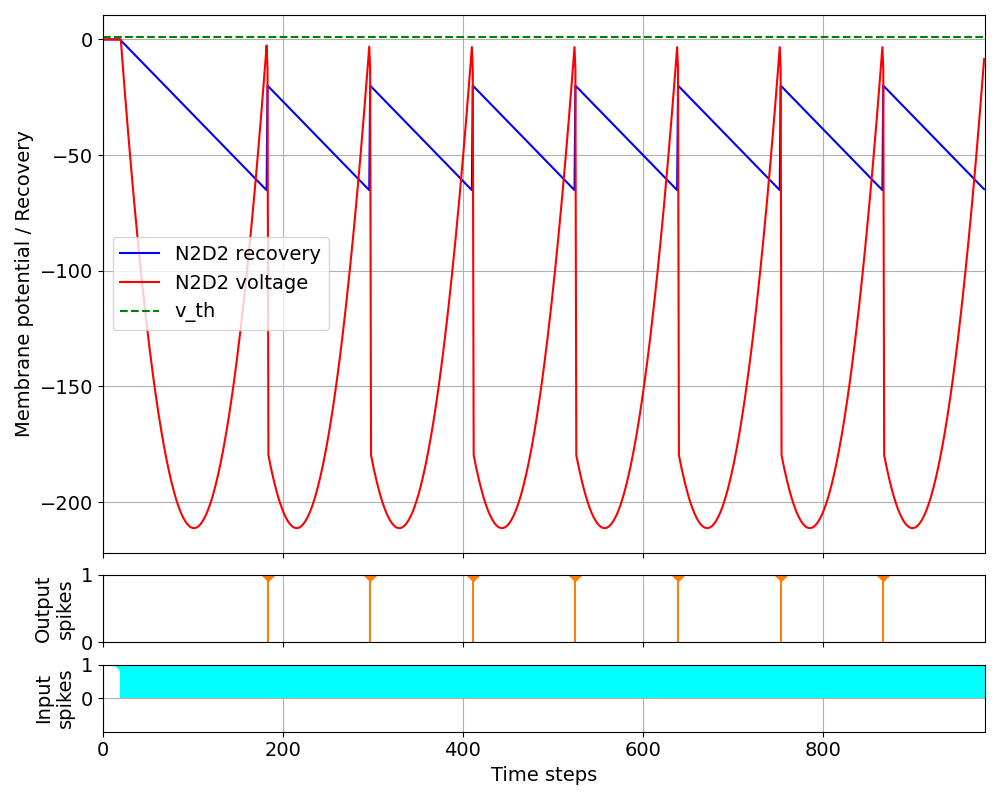}
    \caption{Tonic spiking automatically generated with the optimization framework from the N2D2 model.}
    \label{fig:n2d2_tonic}
\end{figure}

We used the proposed framework to assess the bioplausibility of four neuron models: \ac{lif}, \ac{adex}, Izhikevich, and N2D2. The implementation code and the model parameters corresponding to the various firing patterns, all found by optimizing the corresponding objective, are available in our Git repository. We identified the CmaEsSampler as the most efficient, although the improvements were not substantial compared to  RandomSampler or QEA.

The \ac{adex} and Izhikevich models successfully reproduced all twenty firing patterns, following findings reported in the literature \cite{izhikevich2004whichmodel}. We generated all figures in this paper by employing the framework on the \ac{adex} model, except for Figure  \ref{fig:n2d2_tonic}. We used the \ac{lif} model as a negative control to confirm that the optimization framework does not converge to deceptive solutions that superficially satisfy the objective functions without expressing the target firing patterns. Consistent with prior studies \cite{izhikevich2004whichmodel}, we only found tonic spiking, class 1 excitability, and integrator behavior. For the remaining firing patterns, optimization resulted in high objective values, indicating that the desired behaviors were not realized.

To further test the robustness of the framework, we evaluated the N2D2 model — a custom neuron model not previously analyzed in the literature. While the optimization framework proved effective in this case as well, this model highlighted the practical importance of preliminary manual parameter space exploration. The relevant ranges of model parameters can vary by orders of magnitude, making initial insights into their functional roles highly beneficial for avoiding wasted evaluations in non-informative regions of the parameter space. Additionally, when applying the objective functions to new models, we observed that some models exhibit systematic biases toward certain undesired behaviors. In such cases, minor adjustments to the objective functions may be necessary to guide optimization more effectively. Despite these challenges, the framework successfully identified 11 distinct firing patterns in the N2D2 model without any prior analytical characterization. Figure \ref{fig:n2d2_tonic} shows as an example tonic spiking as found by our optimization framework.
\section{Discussion}
\label{sec:discussion}

This paper demonstrates the viability of automatically identifying neuronal firing patterns by formulating and optimizing appropriate objective functions. However, the implemented objective functions and optimization algorithms are not optimal, and more efficient methods may enable the discovery of patterns with greater speed and robustness.

A key advantage of the proposed optimization framework lies in its modular architecture, facilitated by Optuna \cite{optuna_2019}. This allows new optimization algorithms or objective functions to be implemented and tested independently, without affecting the rest of the framework. The existing objective functions can even be extended to new functionalities. For example, the objective for tonic spiking can be specialized to match a predefined firing frequency — a consideration that can be relevant for \ac{snn} design. Additionally, the Optuna backend naturally supports parallelization of optimization trials when required, allowing for more efficient parameter space exploration.

In terms of objective functions, the main goal was to formulate criteria whose global minimum corresponds to the desired firing pattern. While the implemented objective functions successfully guide optimization toward appropriate solutions, their design could be improved, particularly outside the correct parameter regimes. Currently, many objective landscapes are non-convex and predominantly flat across large regions of the parameter space. The non-convexity largely reflects the complex responses of neuronal models and is, therefore, hard to eliminate. However, the flatness of the objective landscape arises from manually conditioning the objectives to produce large penalties for clearly incompatible cases (such as a neuron firing without stimulus), rather than reflecting a gradual deterioration of performance. With large parameter spaces, these flat regions can become dominant and undermine the search process. An initial, manual exploration of the model helps constrain the search to more plausible regions. Future work should focus on designing objective functions that handle invalid neuronal responses more dynamically, thereby improving the robustness and efficiency of the bioplausibility assessment.

Performance could also be improved by employing more sophisticated optimization algorithms. The CmaEsSampler performed only slightly better than the RandomSampler, and the custom QEA was overly simplified, limiting its potential. Nonetheless, QEAs have demonstrated promising performance in combinatorial optimization literature \cite{han2002quantum,zhang2011quantum}. A related avenue for future research comes from quantum neural network optimization. There, barren plateaus (exponentially flat landscapes) undermine the performance of gradient-based methods \cite{mcclean2018barren}. Specialized algorithms like COBYLA \cite{singh2023benchmarkingquantum} are routinely used to navigate these, which may prove effective for our optimization framework for bioplausibility assessment as well.

As outlined in Section~\ref{sec:introduction}, defining and even more so quantifying the realism of models of biological neural networks remains a complex challenge. A comprehensive assessment would require exact and measurable criteria spanning all levels of neural organization, from large-scale network structure down to the dynamics of synapses, dendrites, and neuronal cell bodies. 
However, given the current availability of quantifiable metrics, neuron models may be a practical entry point for exploring the implications of biological plausibility. Especially in the context of machine learning, it remains an open research question whether biological realism offers tangible advantages, and if so, under which specific conditions or application domains. Moreover, at the level of individual neurons, there is often a trade-off between computational efficiency and strict adherence to biological detail. This renders the level of realism a non-trivial design choice. 
By providing a tool designed to evaluate neuron models in a standardized and accessible manner, treating each model as a black box, we aim to support research in this area. Although our approach does not guarantee to discover all possible firing patterns, it does serve as a viable indicator of a model's general capabilities.
This can help researchers better understand how biological plausibility relates to key performance measures in \acp{snn}, such as prediction accuracy, energy efficiency, robustness, and flexibility. 
In the long term, such an approach may support the development of new neuron models that include only those biological features that clearly improve performance for specific tasks.

\section{Conclusion}
\label{sec:conclusion}

In this work, we presented a new open-source framework for the automated assessment of biological plausibility in spiking neuron models. Our approach builds on the quantitative criteria of the extent to which the firing patterns exhibited by biological neurons can be replicated by the mathematical models. It enables empirical evaluation without requiring detailed prior knowledge or analytical modeling, neuron models are treated as black boxes. The core idea is to encode each of these firing patterns in an objective function and use these to automatically optimize a model's parameters. Our framework successfully recovered all canonical neuronal patterns for well-established models. Furthermore, it demonstrated its utility for characterizing a newly introduced custom neuronal model. While it does not guarantee the discovery of all possible firing behaviors, it provides a flexible and accessible way to gain an overview of a model's capabilities.

The framework is implemented in Python and designed for seamless integration with PyTorch and the \ac{snn} library Norse, making it particularly suitable for machine learning contexts. Our primary goal is to provide a foundation for future research on how biological plausibility affects network-level performance metrics such as accuracy, energy efficiency, robustness, and adaptability. We hope this tool encourages further exploration into the role of biologically inspired design in neuromorphic computing and machine learning.

\begin{credits}

\subsubsection{Author Contributions}
SN is responsible for the idea, core concept and architecture of this work, except the quantum-inspired optimization algorithms. 
AI is responsible for all work related to the quantum-inspired optimization algorithms and supported in the implementation of the framework and the N2D2 neuron. 
AF supported in the implementation of the N2D2 neuron.
All authors contributed to the article and approved the submitted version.

\subsubsection{\ackname}
This research is funded by the German Federal Ministry of Research, Technology and Space
as part of the project ”GreenEdge-FuE“, funding no. 16ME0517K.

\end{credits}

%
%
\bibliographystyle{splncs04}
\bibliography{literature/literature,literature/bioplausible_learning}

\section*{Supplementary}

\begin{table}[]
\centering
    \begin{tabular}{c|c|c|c|c|c}
    Firing pattern & $1/\tau_\text{mem}$ & $1/\tau_\text{syn}$ & $v_\text{leak}$ & $v_\text{th}$ & $v_\text{reset}$ \\
    \hline
    tonic spiking & 60 & 836 & 0 & 18 & 0 \\
    \hline
    class 1 excitability & 100 & 100 & 0 & 18 & 0 \\
    \hline
    integrator & 200 & 200 & 0 & 3 & 0
    \end{tabular}
\caption{LIF neuron parameters}
\label{tab:lif}
\end{table}

\begin{table}[]
\centering
    \begin{tabular}{c|c|c|c|c|c|c|c|c|c}
    Firing pattern & $1/\tau_\text{mem}$ & $1/\tau_\text{syn}$ & $1/\tau_\text{ada}$ & $v_\text{leak}$ & $v_\text{th}$ & $v_\text{reset}$ & $\Delta_\text{T}$ & $a_\text{current}$ & $a_\text{spike}$ \\
    \hline
    tonic spiking & 58.897 & 297.854 & 314.221 & 0 & 27.156 & 0 & 6.225 & 20.775 & 146.281 \\
    \hline
    phasic spiking & 339.493 & 933.016 & 12.903 & 0 & 3.238 & 0 & 2.443 & 5.094 & 143.596 \\
    \hline
    spike frequency adaptation & 4 & 498 & 7 & 0 & 2 & 0 & 11 & 2.3 & 6 \\
    \hline
    tonic bursting & 9.348 & 401.029 & 0.769 & 0 & 3.43 & 29.991 & 5.277 & 49.496 & 61.817 \\
    \hline
    phasic bursting & 1129.476 & 957.405 & -32 & 0 & 68.935 & 0 & 2.342 & 34.658 & 67.024 \\
    \hline
    mixed mode & 190.02 & 64.685 & 1.747 & 0 & 23.597 & 0 & 2.134 & 8.084 & 6.903 \\
    \hline
    class 1 excitability & 514.25 & 668.972 & 141.964 & 0 & 13.138 & 0 & 2.229 & -0.547 & 138.485 \\
    \hline
    class 2 excitability & 1904 & 70 & 7.1 & 0 & 8 & 0 & 8 & 10 & 14 \\
    \hline
    accommodation & 193.744 & 99.84 & 10.396 & 0 & 8.894 & 0 & 9.816 & 7.672 & 18.906 \\
    \hline
    integrator & 96.79 & 79.46 & 100.01 & 0 & 5.79 & 0 & 3.38 & 0.7 & 19.05 \\
    \hline
    spike latency & 102.55 & 115.74 & 14.85 & 0 & 8.22 & 0 & 1.35 & 92.27 & 11.25 \\
    \hline
    depolarizing after-potential & 38.7 & 55.4 & 1.3 & 0 & 0.2 & 0 & 0 & 14.5 & 0 \\
    \hline
    sub-threshold oscillations & 5.299 & 125.91 & 8.598 & 0 & 0.5 & 0 & 6.421 & 19.542 & 5.571 \\
    \hline
    resonator & 314.93 & 94.99 & 12.81 & 0 & 7.73 & 0 & 6 & 49.38 & 11.05 \\
    \hline
    bistability & 20 & 500 & 1 & 0 & 0.1 & -0.1 & 0.2 & 410 & 0 \\
    \hline
    inhibition-induced spiking & 339.074 & 458.859 & -15.451 & 0 & 2.149 & -1.745 & 4.714 & -19.128 & 6.226 \\
    \hline
    inhibition-induced bursting & 397 & 90 & -79 & 0 & 4.5 & -3 & 3.4 & -8 & 0 \\
    \hline
    rebound spiking & 45.231 & 225.073 & 60.781 & 0 & 0.435 & 0 & 1.476 & 21.451 & 3.374 \\
    \hline
    rebound bursting & 82.11 & 477.04 & 76.11 & 0 & 0.32 & 0 & 0.08 & 22.06 & 7.55 \\
    \hline
    threshold variability & 59.844 & 39.82 & 38.146 & 0 & 2 & 0 & 2.637 & 17.51 & 1.194 
    \end{tabular}
\caption{AdEx neuron parameters}
\label{tab:adex}
\end{table}

\begin{table}[]
\centering
    \begin{tabular}{c|c|c|c|c|c|c|c|c}
    Firing pattern & $a$ & $b$ & $c$ & $d$ & $v_\text{rest}$ & $u_\text{rest}$ & $\tau_\text{inv}$ & $v_\text{th}$ \\
    \hline
    tonic spiking & 0.02 & 0.2 & -65.0 & 6.0 & -70.0 & -20.0 & 250.0 & 0.0 \\
    \hline
    phasic spiking & 0.02 & 0.25 & -65.0 & 6.0 & -70.0 & -70.0 & 250.0 & 30.0 \\
    \hline
    spike frequency adaptation & 0.01 & 0.2 & -65.0 & 8.0 & -70.0 & -70.0 & 250.0 & 30.0 \\
    \hline
    tonic bursting & 0.02 & 0.2 & -50.0 & 2.0 & -70.0 & -70.0 & 250.0 & 30.0 \\
    \hline
    phasic bursting & 0.02 & 0.25 & -55.0 & 0.05 & -70.0 & -70.0 & 250.0 & 30.0 \\
    \hline
    mixed mode & 0.02 & 0.2 & -55.0 & 4.0 & -70.0 & -70.0 & 250.0 & 30.0 \\
    \hline
    class 1 excitability & 0.02 & 0.26 & -65.0 & 0.0 & -70.0 & -70.0 & 250.0 & 30.0 \\
    \hline
    class 2 excitability & 0.2 & 0.3 & -70.0 & 0.0 & -70.0 & 0.0 & 250.0 & 30.0 \\
    \hline
    accommodation & 0.002 & 0.6 & -60.0 & 4.0 & -70.0 & 100.0 & 500.0 & -75.0 \\
    \hline
    integrator & 0.02 & -0.1 & -55.0 & 6.0 & -70.0 & -70.0 & 600.0 & 30.0 \\
    \hline
    spike latency & 0.01 & 0.25 & -65.0 & 6.0 & -70.0 & -70.0 & 250.0 & 30.0 \\
    \hline
    depolarizing after-potential & 1.0 & 0.2 & -60.0 & -21.0 & -70.0 & -70.0 & 500.0 & 30.0 \\
    \hline
    sub-threshold oscillations & 0.08 & 0.26 & -60.0 & 0.0 & -70.0 & -70.0 & 250.0 & 30.0 \\
    \hline
    resonator & 0.1 & 0.26 & -60.0 & -1.0 & -70.0 & -63.0 & 500.0 & 30.0 \\
    \hline
    bistability & 1.0 & 1.5 & -72.0 & 0.0 & -70.0 & -70.0 & 500.0 & 30 \\
    \hline
    inhibition-induced spiking & -0.02 & -1.0 & -60.0 & 8.0 & -70.0 & -70.0 & 500.0 & 30.0 \\
    \hline
    inhibition-induced bursting & -0.026 & -1.0 & -45.0 & 8.0 & -70.0 & -70.0 & 500.0 & 30.0 \\
    \hline
    rebound spiking & 0.03 & 0.25 & -60.0 & 4.0 & -70.0 & -70.0 & 600.0 & 30.0 \\
    \hline
    rebound bursting & 0.03 & 0.24 & -54.0 & 0.0 & -70.0 & -70.0 & 600.0 & -50.0 \\
    \hline
    threshold variability & 0.03 & 0.24 & -60.0 & 4.0 & -70.0 & -63.0 & 500.0 & 30.0
    \end{tabular}
\caption{Izhikevich neuron parameters}
\label{tab:izhikevich}
\end{table}
\end{document}